\documentstyle[colap,oe]{article}

\newcommand{\tsnlp}{{\sc tsnlp}}
\newcommand{\tsct}{{\sf tsct}}
\newcommand{\tsdb}{{\sf tsdb}}
\newcommand{\attribute}[1]{{\em #1\/}}
\newcommand{\eg}{e.g.}
\newcommand{\ie}{i.e.}
\newcommand{\viz}{viz.}

\newfont{\itt}{cmitt10}
\newfont{\tinysl}{cmssi8}
\newfont{\tinysf}{cmss8}
\newfont{\hvdemi}{cmssdc10 scaled 650}
\newfont{\largehvdemi}{cmssdc10}
\newfont{\largehvitalic}{cmssi10 scaled 800}

\author{Sabine~Lehmann$^{\spadesuit}$,
        Stephan~Oepen$^{\heartsuit}$\\
        Sylvie~Regnier-Prost$^{\clubsuit}$,
        Klaus~Netter$^{\heartsuit}$,
        Veronika~ Lux$^{\clubsuit}$,
        Judith~Klein$^{\heartsuit}$,\\
        Kirsten~Falkedal$^{\spadesuit}$,
        Frederik~Fouvry$^{\diamondsuit}$,
        Dominique~Estival$^{\spadesuit}$,
        Eva~Dauphin$^{\clubsuit}$,\\
        Herv{\'e}~Compagnion$^{\spadesuit}$,
        Judith~Baur$^{\heartsuit}$,
        Lorna~Balkan$^{\diamondsuit}$,
        Doug~Arnold$^{\diamondsuit}$ \\[1ex]
        \begin{footnotesize}
        \setlength{\tabcolsep}{0.4em}
        \begin{tabular}{@{}cccc@{}}
          \mbox{}$^{\spadesuit}$ISSCO &
          \mbox{}$^{\heartsuit}$DFKI GmbH &
          \mbox{}$^{\diamondsuit}$CL$|$MT Group & 
          \mbox{}$^{\clubsuit}$Aerospatiale France\\
          Universit{\'e} de Gen{\`e}ve &
          CL Department &
          University of Essex &
          Common Research Center \\
          54, route des Acacias &
          Stuhlsatzenhausweg 3 &
          Wivenhoe Park &
          12, rue Pasteur BP 76 \\
          CH 1227 Gen{\`e}ve &
          D 66123 Saarbr{\"u}cken&
          UK Colchester CO4 3SQ &
          F 92152 Suresnes Cedex \\
          $+$41$\,$-$\,$22$\,$-$\,$705$\,$79$\,$33 &
          $+$49$\,$-$\,$681$\,$-$\,$302$\,$52$\,$82 &
          $+$44$\,$-$\,$1206$\,$-$\,$87$\,$20$\,$86 &
          $+$33$\,$-$\,$1$\,$-$\,$4697$\,$30$\,$61 \\
        \end{tabular}\end{footnotesize}}

\title{{\LARGE\sc tsnlp} --- Test Suites for Natural Language Processing}

\begin{document}

\bibliographystyle{oe}

\maketitle

\begin{abstract}
The growing language technology industry needs measurement
tools to allow researchers, engineers, managers, and customers
to track development, evaluate and assure quality, and assess
suitability for a variety of applications.

The \tsnlp\ (Test Suites for Natural Language Processing)
project\footnote{The project was started in December 1993 and completed
in March 1996; the consortium combines strong expertise in machine
translation, evaluation, and natural language processing respectively
and includes Aerospatiale France as an industrial partner. 

Most of the project results (documents, bibliography, test data, and
software) as well as on-line access to the test suite database can be
obtained through the world-wide web from the \tsnlp\ home page
{\tt http://tsnlp.dfki.uni-sb.de/tsnlp/}. 

The \tsnlp\ project was funded within the Linguistic Research
Engineering ({\scriptsize LRE}) programme of the European Commission
({\scriptsize DG XIII}) under research grant {\scriptsize LRE}-62-089.}
has investigated various aspects of the construction,
maintenance and application of systematic test suites as
diagnostic and evaluation tools for NLP applications.\\
The paper summarizes the motivation and main results of
\tsnlp: besides the solid methodological foundation of the
project, \tsnlp\ has produced substantial (\ie\ larger than
any existing general test suites) multi-purpose and multi-user
test suites for three European languages together with a
set of specialized tools that facilitate the construction,
extension, maintenance, retrieval, and customization of the
test data.

As \tsnlp\ results, including the data and technology, are made
publicly available, the project presents a valuable linguistic resource
that has the potential of providing a wide-spread pre-standard
diagnostic and evaluation tool for both developers and users of NLP
applications.
\end{abstract}

\section{Background and Motivation}
\label{background}

Evaluation of NLP applications plays an increasingly important role in
both the academic and industrial NL communities.  Two tools
traditionally used for evaluating and testing NLP systems are {\em
test suites\/} and {\em test corpora\/} respectively.  The two can be
seen as serving complementary purposes (see \newcite{Bal:Net:Arn:94} and
\newcite{Dau:Lux:Reg:95a}):\ in contrast to text corpora, whose main
advantage is that they reflect naturally occurring data, the key
properties of test suites are (i) {\em systematicity}, (ii) {\em
control over data}, (iii) {\em inclusion of negative data}, and (iv)
{\em exhaustivity}.

Among the main motivations for the \tsnlp\ project
were the lack of general guidelines for the test suite construction, of
adequate and comprehensive test material, and of appropriate tools.   
The resulting duplication of effort among test suite developers
obviously leads to a waste of time and resources.  
In addition, one of the main conclusions of a study of existing tests
suites conducted during the first stage of the project
(\newcite{Est:Fal:Bal:94}) was that the reusability of existing test 
suites is severely hampered by their lack of structure and
annotations.  Indeed, despite the pioneering efforts of
\newcite{Fli:Ner:Sag:87} and \newcite{Ner:Net:Dia:93}, most of the existing
test suites were written for some specific system or simply enumerate
a number of interesting examples and, thus, do not meet the demand for
large, systematic, well-documented, highly-structured and annotated
collections of linguistic material, which is now required by a growing
number of NLP applications.  The TSNLP test suite addresses these
demands and provides powerful tools for the construction and
manipulation of the test data.

On the one hand, since every NLP system (whether commercial or under
development) exhibits specific features which make it unique, and every
user (or developer) of an NLP system has specific needs and
requirements, the \tsnlp\ approach is based on the assumption that, in
order to yield informative and interpretable results, any test suite
used for an actual test or evaluation must be {\em specific\/} (at
least to some degree) to the system and the user.  On the other hand,
since testing or evaluating NLP systems is performed for a variety of
purposes, the \tsnlp\ approach is also guided by the need to provide
test material which is easily {\em reusable\/}.  To achieve these two
goals of specificity and reusability, the traditional notion of a test
suite as a monolithic set of test items has been abandoned in favour
of the notion of a database in which test items are stored together
with a rich inventory of associated linguistic and non-linguistic
annotations.

Thus, the test suite database serves as a virtual (or meta) test suite
that provides the means to extract the relevant subset of the test
data suitable for some specific task.
Using the explicit structure of the data and the \tsnlp\ annotations,
the database engine allows searching and retrieving data from the
virtual test suite, thereby creating a concrete test suite instance
according to arbitrary linguistic and extra-linguistic constraints.
Since, additionally, there are tools provided for the maintenance and
extension of the test suite database, the \tsnlp\ virtual test suite
approach is an essential innovation leading the way to a new generation
of highly-structured reusable test suites.

\section{Test Suite Design and Methodology}
\label{methodology}

Based on a survey of existing test suites and an analysis of the
diagnostic and evaluation requirements of both NL technology developers
and users, \tsnlp\ has developed the methodology for the construction
of {\em core test data}, that is, test items reflecting central language
phenomena and that are applicable to a wide range of applications,
including parsers, grammar checkers, and controlled language checkers
(\newcite{Bal:Fou:Arn:96}).

The \tsnlp\ methodology is designed to optimize (i) {\em control over
test data}, (ii) {\em progressivity}, and (iii) {\em systematicity}.
These are necessary qualities for an adequate,
reusable test suite, which are difficult to find in test corpora.  The
methodology also addresses the specific goals of \tsnlp\ to produce
multi-purpose, multi-user, and multilingual test suites.

\paragraph{Control over test data} What makes test suites valuable in
comparison to corpora is that they can focus on specific linguistic
phenomena and that each phenomenon can be presented both in isolation
and controlled combinations in which as many linguistic
parameters as possible are being kept under control.  This is
particularly the case when a phenomenon is illustrated by
systematic variation over the parameters used to describe this phenomenon,
while all other parts of the test items remain constant.

Vocabulary is an aspect of the test data that needs to be controlled. 
\tsnlp\ achieves this by restricting the vocabulary in size as well as 
in domain.  
Categorially and semantically ambiguous words are avoided where
possible and only included when ambiguity is explicitly tested for. 

Additionally, \tsnlp\ attempts to control the interaction of
phenomena by keeping the test items as small as possible. 
Therefore, a number of guidelines for this purpose (such as {\em use
declarative sentences\/} and {\em avoid modifiers and adjuncts}) is
provided.

\paragraph{Progressivity} Progressivity is the principle of starting
from simple test items and increasing their complexity.  In \tsnlp,
this aspect is addressed by requiring that each test item focuses only
on a single phenomenon (or rather subphenomenon or even feature) which
distinguishes it from all other test items.  This principle not only
ensures systematicity during the test data construction but also
allows test data users to apply the test data in a progressive order
resulting the special attribute
\attribute{presupposition} in the phenomena classification.  Thus, the
precise identification of the coverage of a system and of its
deficiencies is rendered easier.

\paragraph{Systematicity}
Systematicity refers to the depth of coverage of a test suite, with
respect to both well-formed and ill-formed items.  Systematicity in
\tsnlp\ is achieved for well-formed items by the explicit
classification of test items according to phenomena and sub-phenomena.
Negative test data permits testing for overgeneration as well as for
coverage.  Ill-formed items are derived from well-formed ones by
systematic variation of the parameters through the
application of one (or more) of four operations, namely:

\begin{itemize}
  \item {\sc replacement} (\eg\ change of person inside a
        verb in subject-verb agreement)\\
        (French) {\em L' ing{\'e}nieur vient.}\\
        (French) {\em *L' ing{\'e}nieur viens.}
  \item {\sc addition} (\eg\ addition of an
        object NP in a sentence with an intransitive verb)\\
        (German) {\em Der Manager arbeitet.}\\
        (German) {\em *Der Manager arbeitet den Vortrag.}
  \item {\sc deletion}
        (\eg\ deletion of an obligatory complement)\\
        (German) {\em Der Manager h{\"a}lt den Vortrag.}\\
        (German) {\em *Der Manager h{\"a}lt.}
  \item {\sc permutation}
        (\eg\ inverting the order of the verb and the object)\\
        (English) {\em He saw the boy.}\\
        (English) {\em *He the boy saw.}\\
\end{itemize}

In general, the systematicity of test data was greatly enhanced through
the use of special-purpose tools in the data construction and
validation process (see section~\ref{technology} below).

\paragraph{Multilinguality}
Multilinguality is achieved in the \tsnlp\ test suites by covering the
same range of phenomena in English, French and German, and adopting
the same classification for these phenomena in the three languages.
Furthermore, the choice of related terminology for the categorial and
structural description contributes to the comparability and consistency
of the test items (see section~\ref{test-data-construction} for
details).

\paragraph{Documentation}
To enhance the usability and extensibility of \tsnlp\ results, a
three-volume user guide is under preparation providing clear
instructions for the assessment of the methodology, test data, and
tools developed.

\section{\tsnlp\ Annotation Schema}
\label{tsnlp-annotation-schema}

Following its survey of existing test suites and guidelines for the
test suite construction, \tsnlp\ designed a detailed
annotation schema for the test data which does not presuppose a
specific linguistic theory (where this exists), a particular evaluation
situation or application type.

Test data and annotations in \tsnlp\ test suites are
organized at four distinct representational levels:

\begin{itemize}
  \item {\bf Core Data} 
        The core of the test data consists in the individual {\em test
        items\/} together with all general, categorial and structural
        information that is independent of a token phenomenon or
        application.  
        Besides the actual input string, annotations at this level include (i)
        bookkeeping and documentation information (author, date, id
        number), (ii) the item format, its length, category and
        well-formedness code, (iii) the (morpho-)syntactic categories
        and string positions of the lexical and phrasal elements
        constituting the test item, and (iv) an (underspecified)
        representation of its functional structure. 
        Encoding a dependency or functor-argument graph rather than a
        phrase structure tree allows generalizations over potentially
        controversial phrase structure configurations and, thus, avoids
        imposing a specific constituent structure but still can be
        mapped onto one. 
  \item {\bf Phenomenon-Related Data}
        Based on a hierarchical classification of linguistic (and
        extra-linguistic) {\em phenomena\/} (e.g.\ verb valency as a
        subtype of general complementation), each phenomenon is
        identified by a phenomenon id and by its supertype(s).   
        Interaction with other phenomena as well as the phenomena which
        must be presupposed are also given (see
        section~\ref{methodology} on progressivity).   
        In addition, the (syntactic) parameters which are relevant for  
        the phenomenon (\eg\  the number and type of
        complements in the case of verb valency) are
        described. 
        Individual test items can be assigned to one or several
        phenomena and annotated according to the corresponding parameters. 
  \item {\bf Test Sets}
        Test items can optionally be grouped into {\em test sets\/}.  
        A test set is a group of test items containing typically one
        positive example and one or more negative examples.  
        The relation between positive and negative test items has been
        one of the most challenging questions in designing test
        data and, as has been mentioned, is based on the systematic
        variation of phenomenon-specific parameters.
  \item {\bf User and Application Parameters}
        Information that typically correlates with the use of a test suite for 
        different types of evaluation and for different
        applications (\eg\ ratings
        of frequency or relevance for a particular domain) is factored
        from the remainder of the data into {\em user \& application
        profiles\/}. 
        As part of the customization process users of the \tsnlp\ test
        suites are encouraged to extend this part of the test suite
        database and add whatever (formal or informal) information is
        necessary for their specific requirements.
\end{itemize}

\begin{figure}
  \begin{center}\begin{footnotesize}
    \setlength{\tabcolsep}{0.2em}
    \begin{tabular*}{7.5cm}{@{}@{\extracolsep{\fill}}|lll|@{}}
      \hline
      \multicolumn{3}{|c|}{{\bf Test Item}}\\
      \hline
      \hline
      {\sf item id:}\ {\em 24020101} & {\sf author:}\ {\em issco}
      & \multicolumn{1}{l|}{{\sf date:}\ {\em jan-95}}\\
      {\sf register:}\ {\em formal}\hspace{1.5em} & {\sf format:}\ {\em none}
      & \multicolumn{1}{l|}{{\sf origin:}\ {\em invented}}\\
      {\sf difficulty:}\ {\em 1} & {\sf wellformedness:}\ {\em 1}
      & \multicolumn{1}{l|}{{\sf category:}\ {\em S}}\\
      \multicolumn{2}{|l}%
                    {{\sf input:}\
                     {\em L' ing\'{e}nieur vient .}}
      & \multicolumn{1}{l|}{{\sf length:}\ {\em 3}}\\
      \multicolumn{3}{|l|}{{\sf comment:}}\\
      \hline
    \end{tabular*}\\
    \begin{tabular*}{7.5cm}{@{}@{\extracolsep{\fill}}|cllcc|@{}}
      \hline
      {\sf position} & {\sf instance} & {\sf category} 
      & {\sf function} & {\sf domain}\\
      \hline
      {\em 0:2} & {\em L' ing\'{e}nieur} & {\em NP\_sg}
      & {\em subj} & {\em 2:3}\\
      {\em 2:3} & {\em vient} & {\em V\_3-sg}
      & {\em func} & {\em 0:3}\\
      \hline
    \end{tabular*}
    \\[0.5ex]
    \begin{tabular*}{7.5cm}{@{}@{\extracolsep{\fill}}|lll|@{}}
      \hline
      \multicolumn{3}{|c|}{{\bf Phenomenon}}\\
      \hline
      \hline
      {\sf phenomenon id:}\ {\em 2402} & {\sf author:}\ {\em issco}
      & \multicolumn{1}{l|}{{\sf date:}\ {\em jan-95}}\\
      \multicolumn{3}{|l|}{{\sf name:}\ {\em
      C\_Complementation\_subj(NP)\_V}}\\
      \multicolumn{3}{|l|}{{\sf supertypes:}\ {\em C\_Complementation}}\\
      \multicolumn{3}{|l|}{{\sf presupposition:}\
                           {\em C\_Agreement, NP\_Agreement}}\\
      {\sf restrictions:}\ {\em neutral} & {\sf interaction:}\ {\em none} 
      & \multicolumn{1}{l|}{{\sf purpose:}\ {\em test}} \\
      \multicolumn{3}{|l|}{{\sf comment:}\ {\em Intransitive verb
      (valency:1)} }\\
      \hline
    \end{tabular*}
  \end{footnotesize}\end{center}
  \caption{Sample instance of the \tsnlp\ annotation schema for one
           test item:\
           the annotations are given in tabular form for the {\em test
           item}, {\em analysis}, and {\em phenomenon\/} levels.}
  \label{annotation-schema-example}
\end{figure}

In addition to the parts of the annotation schema that follow a
formal specification, there is room for textual comments at the
various levels to accommodate information that cannot or need not be
formalized.

\section{Test Data Construction}
\label{test-data-construction}

Following the \tsnlp\ test suite guidelines (\newcite{Est:Fal:Bal:94})
and using the annotation schema sketched above, the construction of
test data was based on a classification of the (syntactic) phenomena to be
covered.  From judgements on the linguistic relevance and frequency for
the individual languages, the following list of {\em core phenomena\/} for
\tsnlp\ was compiled:
\begin{itemize}
  \item complementation;
  \item agreement;
  \item modification;
  \item diathesis;
  \item modality, tense, and aspect;
  \item sentence and clause types;
  \item word order;
  \item coordination;
  \item negation; and
  \item extragrammatical (\eg\ parentheticals and temporal expressions).
\end{itemize}

\begin{figure}
  \begin{center}\begin{footnotesize}
  \setlength{\tabcolsep}{0.18em}
  \begin{tinysf}\begin{tabular}%
        {@{}|l|r@{$|$}l|r@{$|$}l|r@{$|$}l|@{}}
    \hline
    {\bf Phenomenon} & \multicolumn{2}{|c|}{{\bf English}} 
    & \multicolumn{2}{|c|}{{\bf French}} 
    & \multicolumn{2}{|c|}{{\bf German}}\\
    \hline
    \hline
    C\_Complementation & 148 & 863 & 188 & 567 
    & 218 & 246\\
    C\_Agreement & 68 & 55 & 104 & 183 
    & 224 & 175\\
    C\_Modification & \multicolumn{2}{|c|}{} & 329 & 63 
    & \multicolumn{2}{|c|}{}\\
    NP\_Complementation & 10 & 27 & 12 & 28 
    & \multicolumn{2}{|c|}{}\\
    NP\_Agreement & 201 & 995 & 272 & 1082 
    & 299 & 1732\\
    NP\_Modification & 301 & 484 & \multicolumn{2}{|c|}{}
    & 53 & 60\\
    Diathesis & 157 & 124 & 176 & 119
    & 147 & 148\\
    Tense Aspect Modality & 157 & 39 & 77 & 275 
    & 186 & 134\\
    Sentence Types & 80 & 100 & 389 & 387 
    & 105 & 14\\
    Coordination & 147 & 186 & 379 & 319 
    & 105 & 429\\
    Negation & 289 & 129 & 68 & 100
    & 82 & 210\\
    Word Order & \multicolumn{2}{|c|}{} & 7 & 7 & 60 & 160\\
    Extragrammatical & 24 & 34 & \multicolumn{2}{|c|}{} & 253 & 0\\
    \hline
    \hline
    {\bf Total} & 1582 & 3036 & 2001 & 3130 
    & 1732 & 3308\\
   \hline
  \end{tabular}\end{tinysf}
  \end{footnotesize}\end{center}
  \caption{Status of the \tsnlp\ data (December 1995):\ relevance
           and breadth of individual phenomena present
           language-specific variation (the numbers given are for grammatical
           vs.\ ungrammatical items).
           The cross-classification of phenomenon names results from attaching
           the syntactic domain as a prefix to the phenomenon name
           (\eg\ {\em C\_Complementation, NP\_Agreement\/} et al.).
	   Individual phenomena are often further sub-classified
           according to phenomenon-internal dimensions.}  
  \label{status-of-test-data-construction}
\end{figure}

A further sub-classification of phenomena is made according to the relevant {\em
syntactic domains\/} in which a phenomenon occurs (\eg\ sentences
({\small S}), clauses ({\small C}), noun phrases 
({\small NP}) et al.).
Figure \ref{status-of-test-data-construction} gives an overview of the
test material available.
For each of the three languages some 5000 test items are provided.
Therefore, \tsnlp\ has already achieved a substantially broader and
deeper coverage than previous general-purpose test suites (the still
very popular Hewlett-Packard test suite, for instance, has a coverage
of 3000 test items for English only).

In order to enforce consistency of annotations across the
three languages, canonical lists of the categories and
functions used in the description of categorial and
dependency structure were established.  The dimensions
chosen in the classification attempt to avoid the
presupposition of very specific assumptions of a particular
theory of grammar (or of a language), and rather try to
capture those distinctions that seem to be relevant
across the set of \tsnlp\ core phenomena.

\section{Test Suite Technology}
\label{technology}

Because the test data construction proper as well as the customization
and application of a general-purpose test suite to a specific NLP
system or domain are laborious, cost-intensive and error-prone tasks,
\tsnlp\ put strong emphasis on supplying suitable special-purpose tools
to facilitate both the development as well as usage of the \tsnlp\ test
data (\newcite{Oep:Fou:Net:96} give an overview).

\subsection{Test Data Construction}

To ease the time-consuming test data construction and to reduce erratic
variations in filling in the \tsnlp\ annotation schema, a graphical
test suite construction tool (\tsct) was implemented.
The tool instantiates the annotation schema (see
section~\ref{tsnlp-annotation-schema}) as a form-based input mask
and provides for (limited) consistency checking of the field values.
Additionally, \tsct\ allows reusing previously constructed and
annotated data, as quite often --- when constructing a series of test
items --- it can be easier to duplicate and adapt a similar item
rather than produce annotations from scratch.

Additionally, for some of the test data a DCG-based test suite
generation tool (\newcite{Arn:Ron:Fou:94}) was deployed to automatically
produce systematically varied (\ie\ both grammatical and ungrammatical)
test items together with some part of the annotations.

\subsection{Test Data Maintenance and Retrieval}

\begin{figure}
  \setlength{\unitlength}{0.35mm}
  \begin{center}
    \begin{picture}(220,110)(3,0)
      \thicklines
      \put(65,70){\framebox(30,30){\shortstack{\hvdemi ASCII\\[0.6ex]
                                               \hvdemi Query\\[0.6ex]
                                               \hvdemi Interpreter}}}
      \put(105,70){\framebox(30,30){\shortstack{\hvdemi Network\\[0.6ex]
                                                \hvdemi Database\\[0.6ex]
                                                \hvdemi Server}}}
      \put(185,70){\dashbox{5}(30,30){\shortstack{\hvdemi Graphical\\[0.6ex]
                                                \hvdemi Browser\\[0.6ex]
                                                \hvdemi Tool}}}
      \put(160,85){\makebox(0,0){\LARGE\bf\ldots}}
      \put(77.5,69){\vector(0,-1){7.5}}
      \put(82.5,61.5){\vector(0,1){7.5}}
      \put(117.5,69){\vector(0,-1){7.5}}
      \put(122.5,61.5){\vector(0,1){7.5}}
      \put(197.5,69){\vector(0,-1){7.5}}
      \put(202.5,61.5){\vector(0,1){7.5}}
      \multiput(2.5,65)(15,0){4}{\line(1,0){10}}

      \put(75,50){\framebox(130,10){\largehvdemi Library
                                    of Interface Functions}}
      \put(75,30){\framebox(130,10){\largehvdemi Database Inference Engine}}
      \put(115,49){\vector(0,-1){8}}
      \put(165,41){\vector(0,1){8}}
      \multiput(2.5,25)(15,0){4}{\line(1,0){10}}
      \put(82.5,29){\vector(0,-1){8}}
      \put(87.5,21){\vector(0,1){8}}
      \put(137.5,29){\vector(0,-1){8}}
      \put(142.5,21){\vector(0,1){8}}
      \put(192.5,29){\vector(0,-1){8}}
      \put(197.5,21){\vector(0,1){8}}

      \thinlines
      \put(60,0){\line(1,0){40}}
      \put(70,20){\line(1,0){40}}
      \put(60,0){\line(1,2){10}}
      \put(100,0){\line(1,2){10}}
      \put(61,0){\line(1,2){10}}
      \put(99,0){\line(1,2){10}}
      \put(82,10){\makebox(0,0){\shortstack{\largehvitalic\hspace{1.3ex}
                                            English\\[0.2ex]
                                            \largehvitalic Test Data}}}

      \put(115,0){\line(1,0){40}}
      \put(125,20){\line(1,0){40}}
      \put(115,0){\line(1,2){10}}
      \put(155,0){\line(1,2){10}}
      \put(116,0){\line(1,2){10}}
      \put(154,0){\line(1,2){10}}
      \put(137,10){\makebox(0,0){\shortstack{\largehvitalic\hspace{1.3ex}
                                             French\\[0.2ex]
                                             \largehvitalic Test Data}}}

      \put(170,0){\line(1,0){40}}
      \put(180,20){\line(1,0){40}}
      \put(170,0){\line(1,2){10}}
      \put(210,0){\line(1,2){10}}
      \put(171,0){\line(1,2){10}}
      \put(209,0){\line(1,2){10}}
      \put(192,10){\makebox(0,0){\shortstack{\largehvitalic\hspace{1.3ex}
                                             German\\[0.2ex]
                                             \largehvitalic Test Data}}}

      \put(25,10){\makebox(0,0){\footnotesize\sc Data Files}}
      \put(25,45){\makebox(0,0){\shortstack{\footnotesize\sc Database\\[0.3ex]
                                            \footnotesize\sc Kernel\\[0.3ex]
                                            \footnotesize\sc (Server)}}}
      \put(25,85){\makebox(0,0){\shortstack{\footnotesize\sc Client \&\\[0.3ex]
                                            \footnotesize\sc Application\\[0.3ex]
                                            \footnotesize\sc Programs}}}
    \end{picture}
  \end{center}
  \caption{Sketch of the modular \tsdb$_1$ design:\ the database
           kernel is separated from client programs through a layer of
           interface functions.}
  \label{tsdb-architecture}
\end{figure}
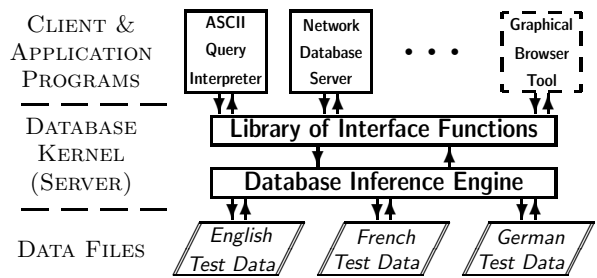

To implement the \tsnlp\ virtual test suite approach (see
section~\ref{background}), the test data is mounted on a relational
database to satisfy the following key desiderata:
\begin{itemize}
  \item {\bf usability}: to facilitate the application of the
        methodology, technology, and test data developed in \tsnlp\ to
        a wide variety of diagnosis and evaluation purposes for
        different applications by developers or users with varied
        backgrounds;
  \item {\bf suitability}: to meet the specific necessities of
        storing and maintaining natural language test data (\eg\ in
        string processing) and to provide maximally flexible interfaces;
  \item {\bf adaptability and extensibility}: to enable and encourage
        users of the database to add test data and annotations
        according to their needs without changes to the underlying data
        model; and
  \item {\bf portability and simplicity}: to make the results of
        \tsnlp\ available on several different hard- and software
        platforms and easy to use.
\end{itemize}

To account for the potentially different requirements of NLP
developers and users and in order to provide suitable interfaces to
human test suite users as well as to external application programs, a
dual database implementation was carried out:\ (i) while a proprietary
implementation (called \tsdb$_{\mbox{\small\sf 1}}$) allowed the
fine-tuning of both the query language and interfaces, (ii) a second version
(\tsdb$_{\mbox{\small\sf 2}}$) builds on a commercial database product
and, thus, is compliant to common industry standards allowing
(industrial) users of the \tsnlp\ test suite to acquire on-site
technical support where necessary.

The \tsdb$_{\mbox{\small\sf 1}}$ implementation is a small and
efficient relational database engine in ANSI C. 
It was designed with an open and documented interface layer (see
figure~\ref{tsdb-architecture}) that enables test suite users to
bidirectionally link an application being tested to the database and
run automated retrieve, process, and compare cycles.
Diagnostic results obtained can be stored into the database as
part of the {\em user \& application profile\/} (see
section~\ref{tsnlp-annotation-schema}) for use in continuous
progress evaluation (section~\ref{customization} gives an example).

An ASCII-based command shell interprets a simplified SQL-style query
language and provides editing, completion, and command and query result
history.
A network database server gives remote (though read-only) access to the
test data.

For the alternative implementation \tsdb$_{\mbox{\small\sf 2}}$ the
competitively priced database package Microsoft FoxPro was deployed
because it is available for both Apple Macintosh and personal computers
running MS Windows\footnote{Building on the popular database package MS
Access, another implementation of the test suite database is currently
being developed.  This version will provide a similar functionality to
\tsdb$_{\mbox{\footnotesize\sf 2}}$ and be available for the MS Windows
world.}
and has a very wide distribution.
The database provides convenient graphical browsing and editing of the
data (using pull-down menus for finite domain fields) as well as
standard import and export facilities to exchange data with external
applications.

\subsection{Query and Retrieval:\ An Example}

To illustrate the capacity and flexibility of the \tsnlp\ annotation
schema in conjunction with a relational database retrieval engine, a
query example in the simplified {\sf SQL}-like query language
interpreted by \tsdb$_{\mbox{\small\sf 1}}$ together with an 
informal English paraphrase is given:\footnote{Additional sample queries and
more details on the database schema (including relation and attribute
names) can be found in \newcite{Oep:Net:Kle:96} and from the \tsnlp\
World-Wide Web home page {\tt http://tsnlp.dfki.uni-sb.de/tsnlp/}.}
\begin{itemize}
  \item find all grammatical test items that are associated with the
        phenomenon of clausal (\ie\ subject verb) agreement and have
        pronominal subjects:
        \begin{tt}\begin{tabbing}
          {\itt se}\= {\itt lect} i-id i-input\\
                   \> {\itt where} \= i-wf $=$ 1 \&\\
                   \>              \> p-name $=$ "C\verb|_|Agreement" \&\\
                   \>              \> a-function $=$ "subj" \&\\
                   \>              \> a-category $\sim$ "\verb|^|PRON"
        \end{tabbing}\end{tt}
\end{itemize}

\section{Customization and Testing}
\label{customization}

To validate the \tsnlp\ annotation methodology, test data, and tools,
the project results have been tested against three different
application types, \viz\ a commercial grammar checker for French, a
controlled language checker ({\small SECC\/}) for English
and a parser (the {\small HPSG} system developed at {\small DFKI}) for
German.
As in this setup the evaluation situations ranged from user-level black
box evaluation of a commercial product to glass
box diagnosis of a research prototype under development (the {\small
DFKI} system), a number of interesting results were
obtained on both the adequacy of the \tsnlp\ approach as well
as the quality of the systems being tested.

\paragraph{French Grammar Checker}

The real life evaluation scenario (\ie\ the diagnostic evaluation
of a commercial NLP product) enabled
Aerospatiale to give a precise account of the type of information
obtainable from the use of \tsnlp.

The following major performance characteristics were revealed:

\begin{itemize}
  \item \tsnlp\ ill-formed test items are frequently not detected
        as such.
  \item The system performs well on (both well-formed and
        ill-formed) test items illustrating the phenomenon of
        agreement, in clauses as well as in noun phrases. 
  \item The system does not master the phenomenon of complementation,
        especially not in adjectival phrases.
  \item Sentential test items produce better results than subsentential
        ones.
  \item The analysis capabilities of the system are limited (19\% of
        the \tsnlp\ test items were not fully analysed). 
\end{itemize}

The interpretation of the results produced by the system and
the comparison with the linguistic information provided in the \tsnlp\
annotations led to an identification of the major shortcomings of
the system in terms of systematicity, lexical and morpho-syntactic
deficiencies, and interference with other system components.

\paragraph{English Controlled Language Checker}

Essex tested the controlled language checker {\small SECC}
(\newcite{Adriaens:94}).  Like Aerospatiale, Essex was mostly in a black
box situation with respect to the system, except that they had access
to the controlled grammar language descriptions (but not to the system
rules).  The testing involved the writing of a large number of
customised test items, due to the fact that many CL rules are
lexically based, whereas the core test suite concentrates on syntactic
phenomena.  The testing proved very valuable in highlighting
deficiencies in the system performance, as well as in the rule
descriptions and gave pointers to the possible source of those errors.

\paragraph{German Parser}

In connecting the German \tsnlp\ test suite to the {\small DFKI}
{\small HPSG} system\footnote{The {\scriptsize DFKI} {\scriptsize HPSG}
system is a state-of-the-art NL core engine and grammar engineering
platform; it is in active use at several research institutions
(including {\scriptsize CSLI} Stanford, Brandeis, Ohio State, and Simon
Fraser Universities), primarily for {\scriptsize HPSG}-style grammar
development for German, English, Japanese, and Italian.}
both the test data as well as the \tsnlp\ technology were validated.
Building on the C version of the \tsnlp\ database
(\tsdb$_{\mbox{\small\sf 1}}$), a bidirectional interface to the
application was established allowing the instantiation of a {\small
DFKI} user \& application profile for the storage of
application-specific data (including performance measures and a
semantic specification of the expected output).

The seamless coupling between the test suite and the NL system allows
running fully automated {\em retrieve, process, and compare\/} cycles
in the continuous progress evaluation of the grammar and software such
that --- after making changes to the system --- the impact on coverage
and performance can be determined in an overnight batch job.
The \tsnlp\ test data and database technology proved to be a highly
adequate tool for glass-box diagnostic evaluation; besides, the
testing experience provided valuable feedback for both the test
suite and the application tested (\newcite{Dau:Lux:Reg:95b}).

\section{Conclusion and Future Work}

The \tsnlp\ project has laid the foundations for building large scale
reference data for diagnostic and evaluation purposes. 
The project has produced a substantial set of test items for three
different languages, which are based on a systematic and
controlled methodology, comprehensively annotated, and 
embedded in an environment allowing for easy access and maintenance
of the data. 
The approach has been successfully tested against commercial and
research NLP applications and components.

However, while this work can be seen as an important step in the right
direction, we are very well aware of future developments which will be
essential for a widespread acceptance of the system in a broad user
community. 
These developments comprise amongst others further extensions of the 
test data (possibly taking into account aspects of morphology and
discourse), customization tools, which support the adaptation of the 
test data to specific domains and applications, as well as tools and
methods which relate the isolated test items to corpora in order to
determine their frequency and relevance.
While the members of the project will continue this work, outside
developers and users of NLP applications are invited to contribute to
these resources which can become a reference standard only if they are
truly public domain.  

\section*{Acknowledgements}

In its initial specification and in the early phase of the
project, \tsnlp\ was greatly inspired by the conceptional and
administrative contributions of Siety Meijer of University of Essex.
Additionally, substantial parts of the implementation work at
{\small DFKI} and the University of Essex have been carried out by
diligent research assistants; the authors gratefully appreciate
the enthusiasm of Tom Fettig, Fred Oberhauser, and Martin
Rondell.  
We especially want to thank Roger Havenith, the \tsnlp\
project officer at {\small DG XIII}, for his help throughout the
project and the two external reviewers, Dan Flickinger and John
Nerbonne, for their constructive comments and suggestions.

\begin{footnotesize}

\end{footnotesize}
\end{document}